\documentclass[letterpaper, 10 pt, conference]{IEEEtran}  

\IEEEoverridecommandlockouts                              

\usepackage{graphics} 
\usepackage{epsfig} 
\usepackage{mathptmx} 
\usepackage{times} 
\usepackage{amsmath} 
\usepackage{amssymb}  

\usepackage[T1]{fontenc}
\usepackage{amsfonts}
\usepackage{booktabs}
\usepackage{siunitx}
\usepackage{caption} 
\usepackage{booktabs}
\usepackage{graphicx}
\usepackage{algorithm, algpseudocode}
\usepackage{float}
\usepackage[percent]{overpic}
\usepackage{caption}
\DeclareGraphicsExtensions{.pdf,.png}

\usepackage[font=small,skip=0pt]{subcaption}

\usepackage{array, tabularx, makecell, booktabs}%

\usepackage{geometry}
\usepackage{comment}

\title{\LARGE \bf
Human-Following and -guiding in Crowded Environments using Semantic Deep-Reinforcement-Learning for Mobile Service Robots
}

\author{Linh K{\"a}stner$^{1}$\thanks{$^{1}$Linh K{\"a}stner, Bassel Fatloun, Zhengcheng Shen, Daniel Gawrisch, and Jens Lambrecht are with the Chair Industry Grade Networks and Clouds, Faculty of Electrical Engineering, and Computer Science,				
		Berlin Institute of Technology, Berlin, Germany
		{\tt\small linhdoan@tu-berlin.de}}, Bassel Fatloun$^{1}$, Zhengcheng Shen$^{1}$, Daniel Gawrisch$^{1}$, and Jens Lambrecht$^{1}$
}

\begin{document}

\maketitle
\thispagestyle{empty}
\pagestyle{empty}


\begin{abstract}

Assistance robots have gained widespread attention in various industries such as logistics and human assistance. The tasks of guiding or following a human in a crowded environment such as airports or train stations to carry weight or goods is still an open problem. In these use cases, the robot is not only required to intelligently interact with humans, but also to navigate safely among crowds. Thus, especially highly dynamic environments pose a grand challenge due to the volatile behavior patterns and unpredictable movements of humans. In this paper, we propose a Deep-Reinforcement-Learning-based agent for human-guiding and -following tasks in crowded environments. Therefore, we incorporate semantic information to provide the agent with high-level information like the social states of humans, safety models, and class types. We evaluate our proposed approach against a benchmark approach without semantic information and demonstrated enhanced navigational safety and robustness. Moreover, we demonstrate that the agent could learn to adapt its behavior to humans, which improves the human-robot interaction significantly.

\end{abstract}
\section{Introduction}
\noindent Mobile robots are being increasingly employed for applications such as delivery, health care, and assistance. With the objective of bringing personal robots closer to coexisting with humans in real life, the incorporation of various high-level, human-robot tasks has been an important direction in both research and industry.
For assistance robots in crowded environments, such as train stations or airports, the human-following, and -guiding task is a common problem that is yet to be solved. There, the robot not only has to ensure safe navigation in highly dynamic environments, but also interact intelligently with the human. For instance, by maintaining proximity to the human or waiting for the human in the scenario that they get stuck. This is especially complex in crowded environments due to the unpredictable and volatile behavior of all actors.
Traditional approaches often employ overly conservative and hand-designed constraints, low speed, or limited operation areas, which makes them inflexible and hinders the widespread application of assistance robots in crowded environments.
With the recent advances in sensor technology and computer vision approaches, incorporating semantics into robotics has become an important focus area in research to enable more complex tasks. Thereby, semantics provide the robot with more information about the environment, such as social states, object relationships, or object classes to reason about and make more intelligent decisions to improve human-robot interaction and collaboration.
In recent years, Deep Reinforcement Learning (DRL) has emerged as an end-to-end method that demonstrated superiority for obstacle avoidance in dynamic environments and for learning complex behavior rules. Thus, a variety research publications incorporated DRL to solve high-level tasks such as grasping, navigation or simulation \cite{chen2017socially}, \cite{faust2018prm}, \cite{dugas2020navrep}, \cite{chen2019crowd}.
In this paper, we incorporate semantic knowledge about the environment like the obstacle position and social states as additional information to train an agent that is not only able to execute point-to-point navigation, but also is able to follow or guide a human through crowded environments.
We introduce a set of semantic reward systems to shape the agent's behavior and propose a DRL agent that can learn to achieve these high-level tasks in an end-to-end manner. Finally, we evaluate our approach within the 2D simulation environment arena-rosnav \cite{kastner2021towards} on a variety of different scenarios.

\begin{figure}[]
    \centering
    \includegraphics[width=0.8\linewidth, height=0.76\linewidth]{ 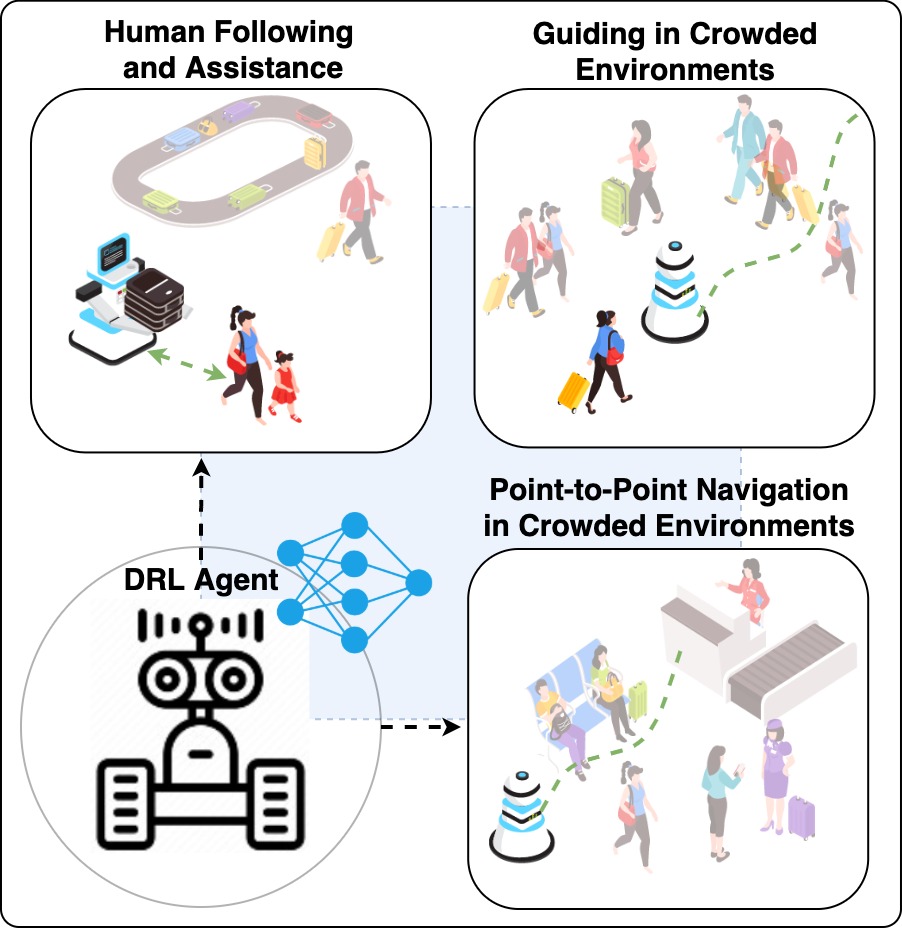}
    \caption{Our proposed agent can solve human-following and -guiding tasks within crowded environments. It was trained with additional semantic information like social states such as human talking, running etc. to reason and perform high level semantic tasks end-to-end.}
    \label{intro}
\end{figure}
The main contributions of this work are the following:
\begin{itemize}
    \item Proposal of a semantic DRL-based agent to navigate safely through crowded environments
    \item Incorporation of semantic information for high-level tasks. More specifically, the agent is able to guide or follow humans in crowded environments.
    \item Qualitative and quantitative evaluation on different highly dynamic environments and comparison against a baseline DRL approach without semantic information

\end{itemize}

\noindent The paper is structured as follows. Sec. II begins with related works. Subsequently, the methodology is presented in Sec. III. Sec. IV presents the results and discussion. Finally, Sec. V will provides a conclusion and outlook. We made the code publicly available under https://github.com/ignc-research/arena-sim-to-real.

\section{Related Works}
\noindent Incorporating semantics into robot navigation has become increasingly important to allow robots to execute more complex tasks, such as the search for specific objects, navigation to certain areas, or following and guiding other actors. Thus, various research works utilize semantics to improve robot performance \cite{ontrup2001hyperbolic}, \cite{yang2018visual}.
Yang et al. \cite{yang2018visual} propose visual semantic navigation using graph convolutional neural networks to incorporate knowledge about objects for a DRL agent. They show how semantic knowledge can benefit navigation performance significantly and is able to help navigate new and unknown scenarios. Other works include Borkowski et al. \cite{borkowski2010towards} or Chaplot et al. \cite{chaplot2020object}, which incorporate semantics for exploration of unknown environments. The researchers demonstrate more efficient exploration by incorporating environmental semantics into a DRL agent.
Works from Chen et al. \cite{chen2021semantic}, \cite{chen2020soundspaces}, \cite{chen2020learning} present audio-visual navigation approaches that incorporate sound into the pipeline to provide the robot with visual and sound cues to improve environmental understanding of the robot.
Drouilly et al. \cite{drouilly2015semantic} propose semantic representations to improve navigation in large-scale environments. The authors employ a semantic map built from spherical images that include high-level information about interactions with humans and demonstrated improved navigation performance in complex outdoor environments.
Besides the basic point-to-point navigation, the human following and human guiding tasks, where a robot should follow or guide the human in crowded and/or unknown situations, are becoming increasingly relevant for service robots in crowded scenarios such as airports or train stations. Thus, a variety of research works have dealt with this problem in the past decades \cite{morioka2004human}-\cite{gupta2016novel}. A majority of the work focuses on the usage of computer vision and data processing approaches to detect a human and guide the robot to follow these detections \cite{hoshino2011human}, \cite{dang2011human}.
Works such as \cite{xiao2017human}, \cite{geetha2021follow}, or \cite{chen2019human} utilize sensors and data processing approaches to track and follow humans.
Ahn et al. \cite{ahn2018formation} propose a method to predict a set of recommended and safe locations based on the person's location. The researchers demonstrate that using their approach, the robot can follow the human more accurately and safely.
Learning-based solutions to the human following task include Algabri et al. \cite{algabri2020deep}, who use a deep learning-based color feature mapping to follow the human.
Chi et al. \cite{chi2017gait} propose a human following approach employing a gait cycle recognition method. Therefore, the researchers designed a segmentation model that predicts different walking patterns and directions from RGB-D data.
A main limitation of the aforementioned works is that the researchers only tested their approaches in simple and predefined environments. However, in most real-world scenarios, the human following or guiding task is coupled with navigation in highly dynamic environments. Furthermore, the robot must react and interact with the human it follows or guides. For instance, when the human is stuck in a crowd or moves significantly slower than the robot, DRL emerged as an alternative planning approach, which is able to generalize new problem instances and learn complex behavior rules. Thus, various research works incorporated DRL into their navigation systems \cite{dugas2020navrep},  \cite{faust2018prm}, \cite{chen2019crowd}, \cite{chen2017socially}.
In this paper, we propose a DRL-based approach, which is able to learn complex behavior rules based on sensor observations and semantic information.
We incorporate semantic knowledge about obstacles such as the social state using a DRL agent. The agent is able to perform navigation in crowded environments as well as perform high-level tasks, such as human following or guiding through crowded environments based on sensor and semantic observations.

\begin{figure}[]
    \centering
    \includegraphics[width=1\linewidth]{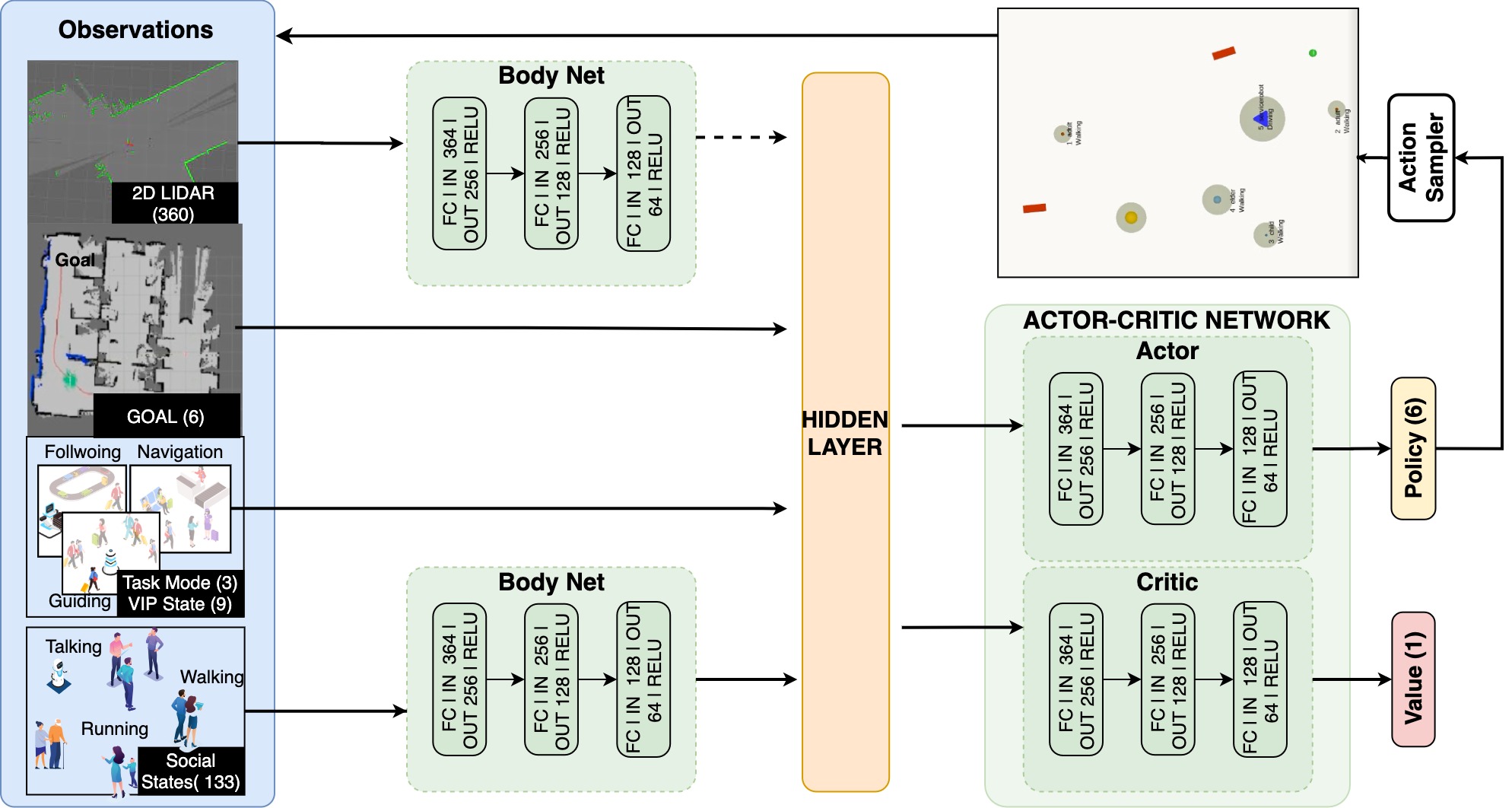}
    \caption{Our proposed agent can solve human following and guiding within crowded environments. It was trained with additional semantic information like social states such as human talking, running etc. to reason and perform high level semantic tasks end to end.}
    \label{system}
\end{figure}

\section{Methodology}
\noindent In this chapter, we present the methodology of our proposed framework.

\subsection{System Design}

\noindent Fig. \ref{system} illustrates the system design of our approach. The DRL agents are trained within our 2D simulation environment arena-rosnav \cite{kastner2020deep}. The observations are processed by the DRL agent, which produces an action in the environment.
Since the arena-rosnav environment originally was designed for point-to-point navigation, we extended it to include two higher-level tasks: the human-following task, and the human-guiding task, where the robot follows or guides a human, denoted as a very important pedestrian (VIP) through a crowded environment. To make the environment more realistic, we integrated the Pedsim project \cite{helbing1995social} into our arena-rosnav, which generates realistic human trajectories based on social states. To trigger our two additional tasks, we added two social states to the Pedsim model: requesting help and request to follow. In the following, these two tasks are explained in more detail.

\subsubsection{Robot Following}
This task consists of two steps: navigating to the pedestrian, taking the guidance service, waiting for the pedestrian to enter the following-guide state, and guiding the human to the final destination while maintaining a small distance from the pedestrian. Algorithm 1 implements the robot-following task.
\begin{algorithm}[h]
\caption{Robot Following Algorithm}\label{alg:cap}

\begin{algorithmic}[1]

\State $i \gets 0$ 
\State $ flag \gets 0$ 
\State $goal  \gets end\_goal$ 
\While{$i \neq len(peds)$}
\If{$peds[i].state == $ "StateRequestingGuide"}
    \State $ flag \gets 1$
    \State $goal  \gets peds[i]$
\ElsIf{$peds[i].state == $ "StateFollowingGuide"  }
\State $ flag \gets 2$
\If{  distance(peds[i],robot) $ > 3.0$ }
 \State $ flag \gets 1$
\State $goal  \gets peds[i]$
\EndIf
\EndIf
\State $i \gets i +1$
\EndWhile

\end{algorithmic}
\end{algorithm}

\subsubsection{Human Following}

This task consists of two stages: navigation towards the offering-guide-service requesting pedestrian, waiting for the pedestrian to enter the guide-to-goal state, successful close-distance following of the pedestrian to the end goal, and waiting for the pedestrian to clear the goal while approaching the end destination carefully. Algorithm 2 describes the implementation of the human following task.

\begin{algorithm}[h]
\caption{Human Following Algorithm}\label{alg:cap}
\begin{algorithmic}
\State $i \gets 0$
\State $ flag \gets 0$
\State $goal  \gets end\_goal$ 
\While{$i \neq len(peds)$}
\If{$peds[i].state == $ "StateRequestingFollower"}
    \State $ flag \gets 3$
    \If{   distance(peds[i],robot) $ > 4.0$ }
\State $goal  \gets peds[i]$
\ElsIf{distance(peds[i],robot)  $\leq  4.0$}
\State $goal  \gets none$
\EndIf
\ElsIf{$peds[i].state == $ "StateGuideToGoal"  }
 \State $ flag \gets 4$
    \If{   distance(peds[i],robot) $ > 4.0$ }
\State $goal  \gets peds[i]$
\ElsIf{distance(peds[i],robot)  $\leq  4.0$}
\State $goal  \gets none$
\EndIf
\ElsIf{$peds[i].state == $ "StateClearingGoal"  }
 \State $ flag \gets 5$
 
\EndIf
\State $i \gets i +1$
\EndWhile
\end{algorithmic}
\end{algorithm}

\subsection{Agent and Neural Network Design}
\noindent In order to complete these tasks, we propose three different agents, each with different observation spaces to study the effect of additional semantic input.
The basic input contains the relative goal position and 2D laser scan data. The number of laser scans is calculated by subtracting the minimum angle from the maximum angle. In cases where this information is irrelevant or even misleading, this field is overwritten with an invalid value such as -1. Next, we defined three agents with additional information as follows:
$SDRL_{SafeZone}$ denotes an agent with additional safety zone information for each obstacle that should make the navigation safer.
The agent without any safety zone information is denoted as $SDRL_{NoSafeZone}$. The agent with both information is denoted as $SDRL_{Complete}$. It receives additional semantic social states as input, which contains relevant information such as type, social status, radius, safety distance, distance to the agent, and position in the agent frame. For simplicity, we only consider the seven nearest pedestrians. If there are no or too few pedestrians, the empty fields are either filled with invalid values or the existing pedestrian observations are duplicated.
To allow the agent to uniquely distinguish its state, we extend the observation by the observations of the last x time steps, where x is a predefined parameter and in our case always equals 8.
Since we intend to teach the agent to distinguish between modes and tasks, we use a numeric flag in the range [0..2], where 0 is the normal navigation task with no special pedestrians to track or guide. The VIP input stands for very important pedestrian and contains information such as the robot's position and speed with respect to the VIP and its orientation, as well as the distance between the robot and the VIP.
In total, the observation collector provides the neural network with a 504 scalar values, which are given as input to the neural network.
The action state of the agent is discrete and contain six actions:  \{forward,stop,left,right,strong left,strong right\}.

\subsubsection{Neural Network Architecture}

For this work, a deep neural network architecture based on Social Attention with Reinforcement Learning (SARL), DNN was chosen.
The time step [T], robot state [S], task flag [F], and the very important pedestrian state [VIP] are directly fed into the value and policy network, which have the same structure. Meanwhile, the lidar scans [L] and the pedestrian information [Ps1. . Psn] pass through a separate body network and continue as hidden states to join the rest of the processed data in the input space [HUMAN] entering the policy and value network. Figure \ref{system} illustrates the network structure as well as in- and output.

\subsection{Reward Functions}
\noindent Since sparse rewards do not lead to fast convergence of the agent, we design our reward function to be dense and return a reward after each transition.
Negative rewards are only given for collisions or if the agent gets too close to a static or dynamic obstacle.
Positive rewards are given when the agent moves toward or reaches the target with a reasonable number of steps: the fewer steps required, the higher the reward. Equation 1 states the reward system of our agents.
\begin{align}
    R_{all}(s_t,a_t) = [r_{g}^t + r_{c}^t + r_{d}^t + r_{ss}^t + r_{sd}^t]
\end{align}

\noindent Where  $r_{s}^t$ is the success reward for reaching the goal, $r_{c}^t$ is the punishment for a collision and both lead to episode ends.
 $r_{d}^t$ describes the reward for approaching the goal. Additionally, we introduce two safety rewards  $r_{ss}^t$ to help avoid static obstacles and $r_{sd}^t$ is meant for dynamic obstacles .\\
\begin{align}
r_{s}^t&=\begin{cases}2&|p_g - p_r|_2 < r_{g}\\0&otherw.\end{cases} \quad\\
r_{c}^t&=\begin{cases}-4& \min(O_{s,t}) < r_r\\
0&otherw.\end{cases} \quad\\
r_{d}^t& =\begin{cases}0.018e^{1-t}& d_{rg}^{t}<d_{rg}^{t-1}\\
-0.03 & \text {else if } d_{rg}^{t}=d_{rg}^{t-1} \\-0.014&  otherw.\end{cases}\quad\\
r_{ss}^t& =\begin{cases}-0.15& d_r < d_{safe}\\ 0&otherw.\end{cases} \quad\\
r_{sd}^t&=\begin{cases}0.08e^{1-d_{sd}} & \text {if exceed } \mathrm{SD}\\0&otherw.\end{cases}
\end{align}

\noindent We define $\min(O_{s,t}) $ as the smallest laser scan distance to agent center, $r_r$ to be the robot radius
, $d_{rg}$ describes the distance between the robot and the goal. Furthermore we set $t=step_{current}/step_{max}$ and  $ d_{sd}=d_{rh}/d_{safe} $. Where $d_{rh}$ is calculated as the distance between the agent and the human and  $d_{safe}$ is a safety distance which adapts to the human types and human behaviors.

\subsubsection{Human Following and Guiding Tasks}
For the human following tasks, the reward r\_d was modified to motivate the agent to adapt to the human's behavior. In the first stage, the granting of a positive reward is conditional on maintaining a minimum distance of 3 meters from the collaborating pedestrian $d_{rp}$ to avoid collisions and ensure comfort. To ensure that the agent does not lose sight of the pedestrian by navigating too quickly to the end destination, an additional condition is imposed in stage 2 that requires the agent to maintain a maximum distance of 4 meters from the pedestrian.
The modified reward function is demonstrated below with $cond_1 $ being  $d_{rp}^{t} \geq 3$ during stage one and $d_{rp}^{t} \leq 4$  during stage two.
 
\begin{align}
r_{d}^t& =\begin{cases}0.018e^{1-t}& d_{rg}^{t}<d_{rg}^{t-1} $ and $  cond_1 \\
-0.03 & \text {else if } d_{rg}^{t}=d_{rg}^{t-1} \\-0.014&  otherw.\end{cases}
\end{align}

\subsection{Training procedure}
\noindent For the human following tasks, to avoid overfitting and achieve a more robust training result, the lead pendant is made to navigate to random targets before finally heading to the final target. In addition, the final state, the clearing-target, is introduced to mark the moment when the agent must start navigating carefully instead of rushing to the target. To make the task easier and more realistic, we also apply a force that pulls the pedestrian away from the robot after reaching the goal to avoid extreme cases, such as running directly into the agent or making sudden turns that block the path to the goal.
We trained the agent in randomized, dynamic environments, where all dynamic obstacles were spawned at random positions and move randomly according to the social force model \cite{helbing1995social}. This way, we mitigate over-fitting issues and enhance generalization. Curriculum training is adapted, which spawns more obstacles once a success threshold is reached and fewer obstacles if the agent's success rate is low. The threshold is defined as the sum over the mean reward.

\begin{figure*}[!h]
    
\begin{subfigure}{0.32\textwidth}(i)
         \centering 
        \includegraphics[width=\linewidth]{ 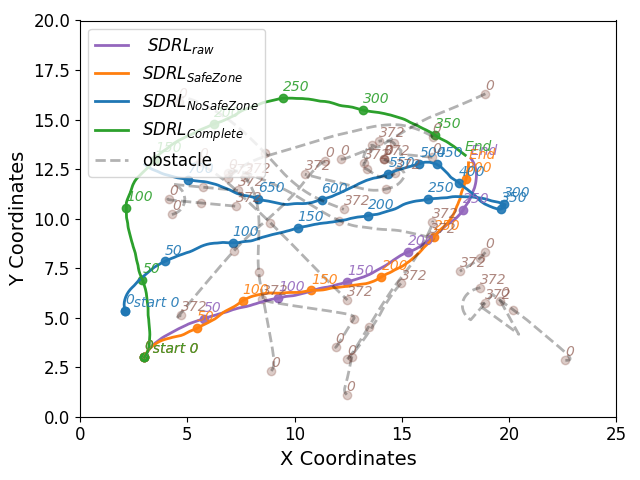}
        
        \label{fig:1}
    \end{subfigure}\hfil 
    \begin{subfigure}{0.32\textwidth }(ii)
     \centering 
        \includegraphics[width=\linewidth]{ 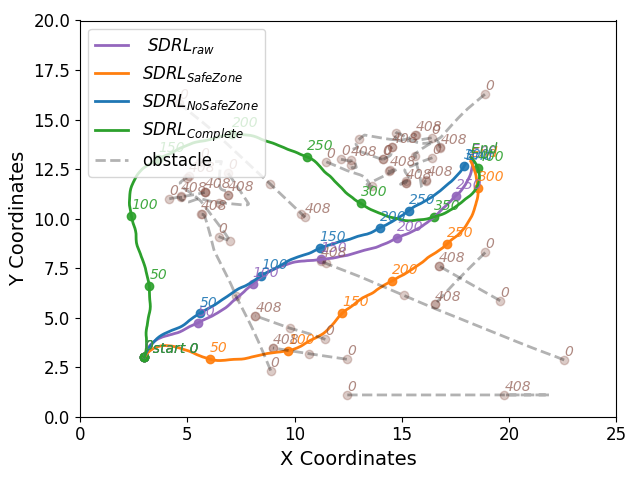}
        
        \label{fig:2}
    \end{subfigure}\hfil 
    \begin{subfigure}{0.32\textwidth}(iii)
     \centering 
        \includegraphics[width=\linewidth]{ 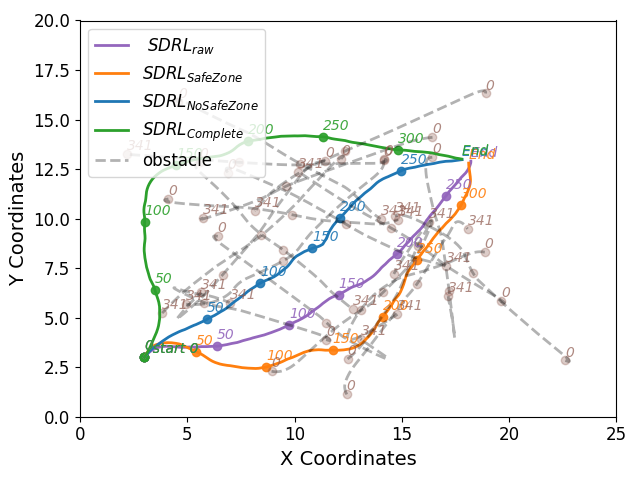}
        
        \label{fig:3}
    \end{subfigure}
    \medskip
    \begin{subfigure}{0.32\textwidth}(a)
         \centering 
        \includegraphics[width=\linewidth]{ 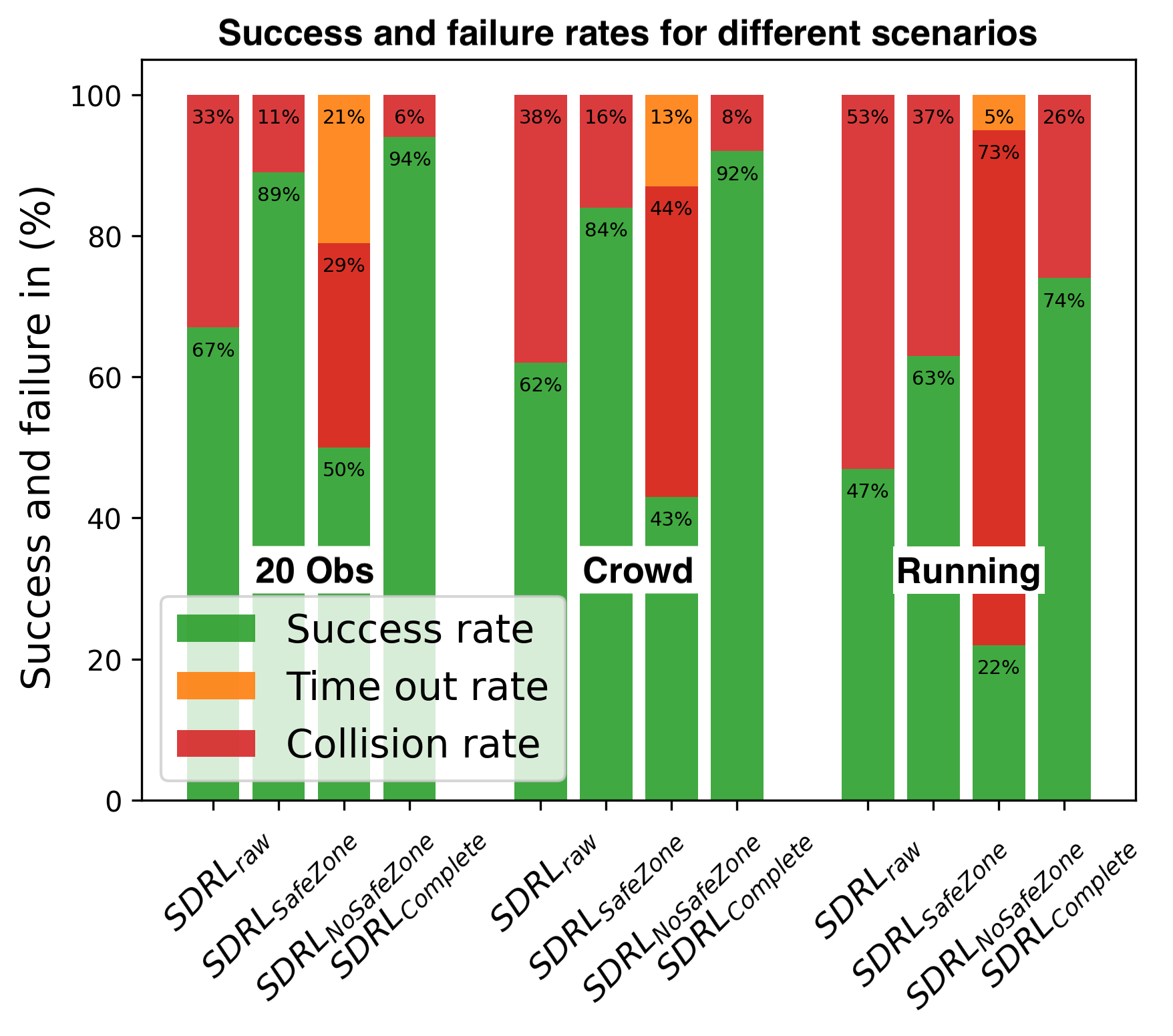}
        
        \label{fig:1}
    \end{subfigure}\hfil 
    \begin{subfigure}{0.32\textwidth }(b)
     \centering 
        \includegraphics[width=\linewidth]{ 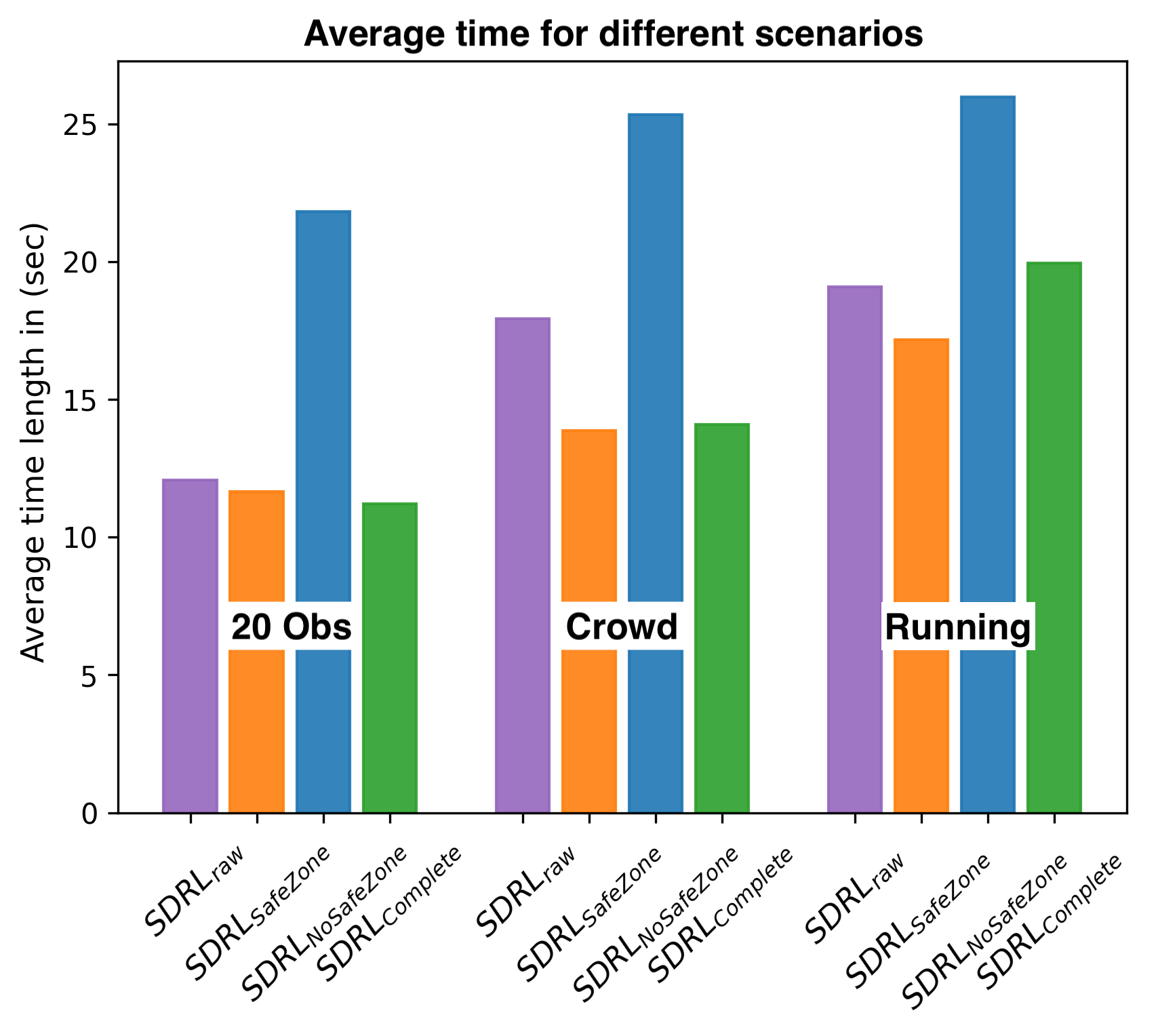}
        
        \label{fig:2}
    \end{subfigure}\hfil 
    \begin{subfigure}{0.32\textwidth}(c)
     \centering 
        \includegraphics[width=\linewidth]{ 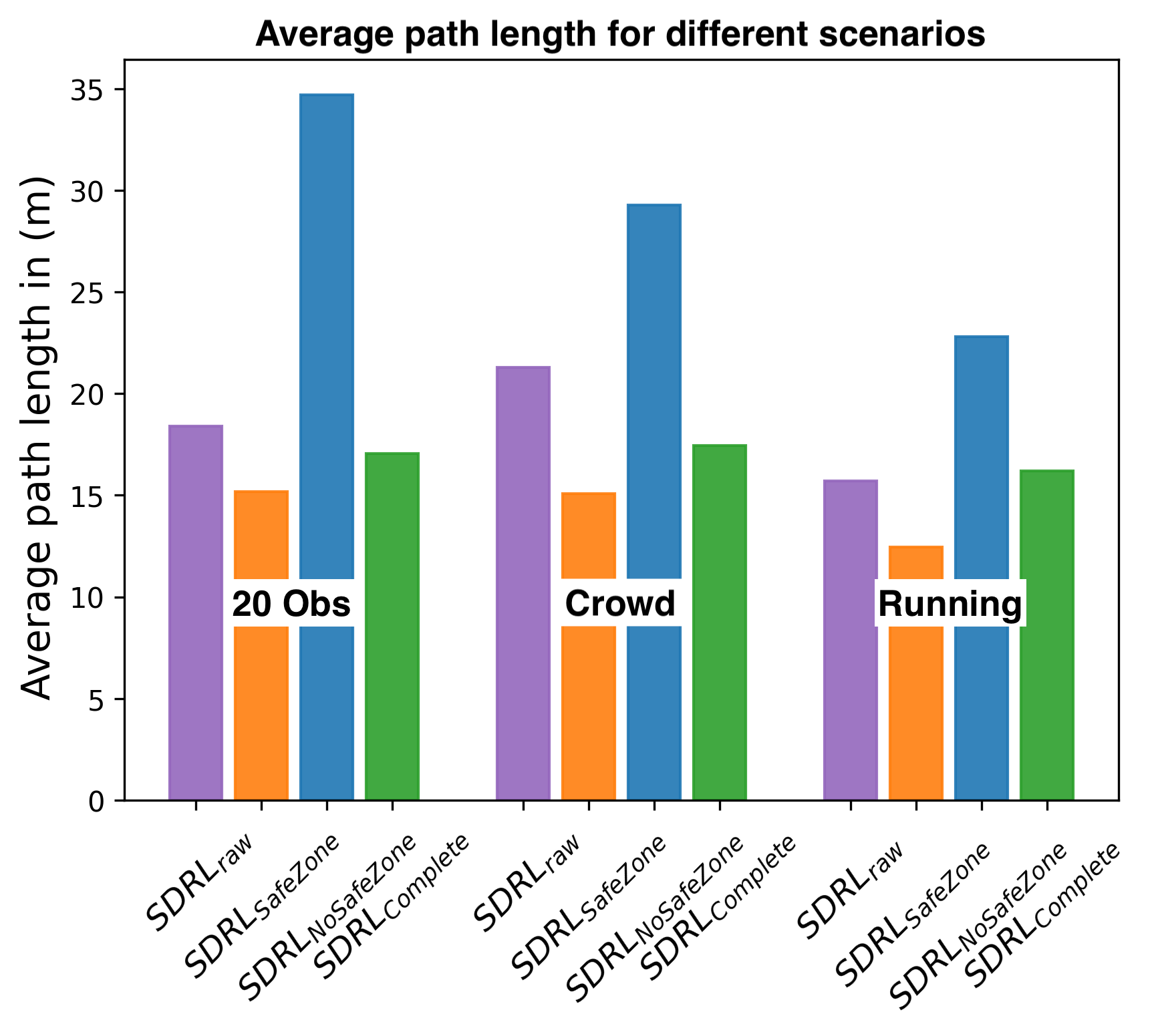}
        
        \label{fig:3}
    \end{subfigure}
    
    \caption{Upper row: trajectories of all agents on three different, highly dynamic scenarios: (i) with 20 obstacles moving randomly, (ii) with obstacles organizing into crowds, and (iii) with running obstacles (up to 1.0 m/s). Lower row: Quantitative evaluations of all agents over 500 episodes: a) the success, failure, and timeout rates, b) the average distance traveled, and c) the average path length. }
    \label{quali}
\end{figure*}

\section{Evaluation}
\noindent In this chapter, we present the evaluation of our semantic agents. The experiments are split into two categories. In the first part, we access the navigational performance of the planners in terms of success rate, efficiency, and robustness.
In the second part, the high-level tasks, human-following and human-guiding are evaluated in crowded environments of up to 20 obstacles. The obstacle velocities are set to 0.3 m/s. The robot's maximum velocity is 0.22. Our semantic agents with the safety distance model $SDRL_{safe}$, without the safety distance model $SDRL_{nosafe}$, and with both safety distances and social states $SDRL_{complete}$ are tested against a baseline DRL agent without any semantic information, which is denoted as $SDRL_{raw}$.

\subsection{Navigational Performance}
\noindent For the qualitative evaluations of the navigational performance, we tested all approaches in three different scenarios: a) with 20 obstacles, b) with obstacle clusters, and c) with running obstacles with an obstacle velocity of up to 1m/s. The scenarios have a fixed start and goal position and the obstacles are moving according to the Pedsim social model \cite{helbing1995social}. The qualitative trajectories of all planners on each scenario are illustrated in the upper row of Fig. \ref{quali}. The timesteps are sampled every 100 ms and visualized within the trajectory of all approaches. The trajectories of the obstacles are marked with the start and end time in seconds. The episode ended once a collision occurred.
It is observed that our semantic DRL approaches avoid all obstacles with a larger margin, whereas the raw DRL approach without semantic information drives with closer proximity to the obstacles.
Our semantic DRL approaches can leverage the additional information about the obstacle type and perform much safer navigation while maintaining efficiency and robustness. This is even more clear in the quantitative results plotted in the lower row of Fig. \ref{quali}.
For the quantitative evaluations, the approaches were statistically tested on random scenarios where the start and goal positions were randomly set and the agent was run 500 episodes each. Subsequently, we calculated the success rate, the average distance traveled, and the average time to reach the goal. The timeout is set to 3 minutes.
It is observed that our agents $SDRL_{complete}$ and $SDRL_{safe}$ accomplish the highest success rates of over 90 and 80 percent on the scenario with 20 obstacles and a crowded environment, and 74 and 63 percent in the running scenario, respectively. Surprisingly, the agent without the safety zone model $SDRL_{nosafe}$ performed worst with under 50 percent success rate on all scenarios while the raw approach accomplished 67 percent success in the first two scenarios, which drops to 47 percent in the running scenario. In terms of efficiency, all planners accomplish similar results in the first scenario with around 12 seconds to reach the goal and 18 meters path length, except for $SDRL_{nosafe}$, which again performed the worst. These results indicate that the agents provided with semantic information are both more safe, robust, and efficient. Furthermore, the semantic agents pay more attention to incoming obstacles and can handle higher obstacle velocity.
The navigation behavior of our planners is demonstrated visually in the supplementary video.
\begin{figure*}[!h]
    
        \begin{subfigure}{0.32\textwidth}(a)
         \centering 
        \includegraphics[width=\linewidth]{ 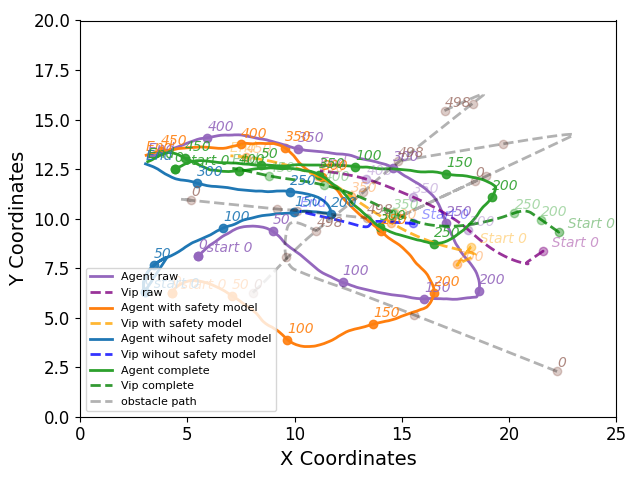}
        
        \label{fig:1}
    \end{subfigure}\hfil 
    \begin{subfigure}{0.32\textwidth }(b)
     \centering 
        \includegraphics[width=\linewidth]{ 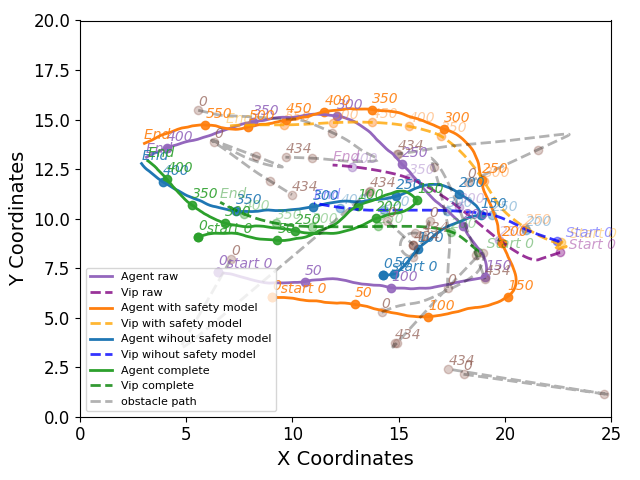}
        
        \label{fig:2}
    \end{subfigure}\hfil 
    \begin{subfigure}{0.32\textwidth}(c)
     \centering 
        \includegraphics[width=\linewidth]{ 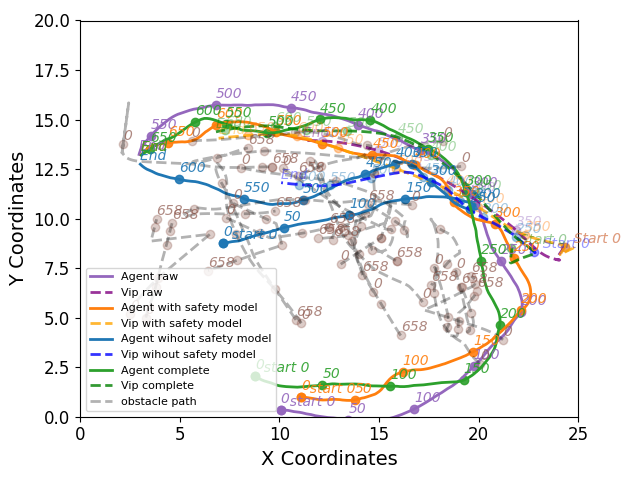}
        
        \label{fig:3}
    \end{subfigure}
    \medskip
    \begin{subfigure}{0.33\textwidth}
         \centering 
        \includegraphics[width=\linewidth]{ 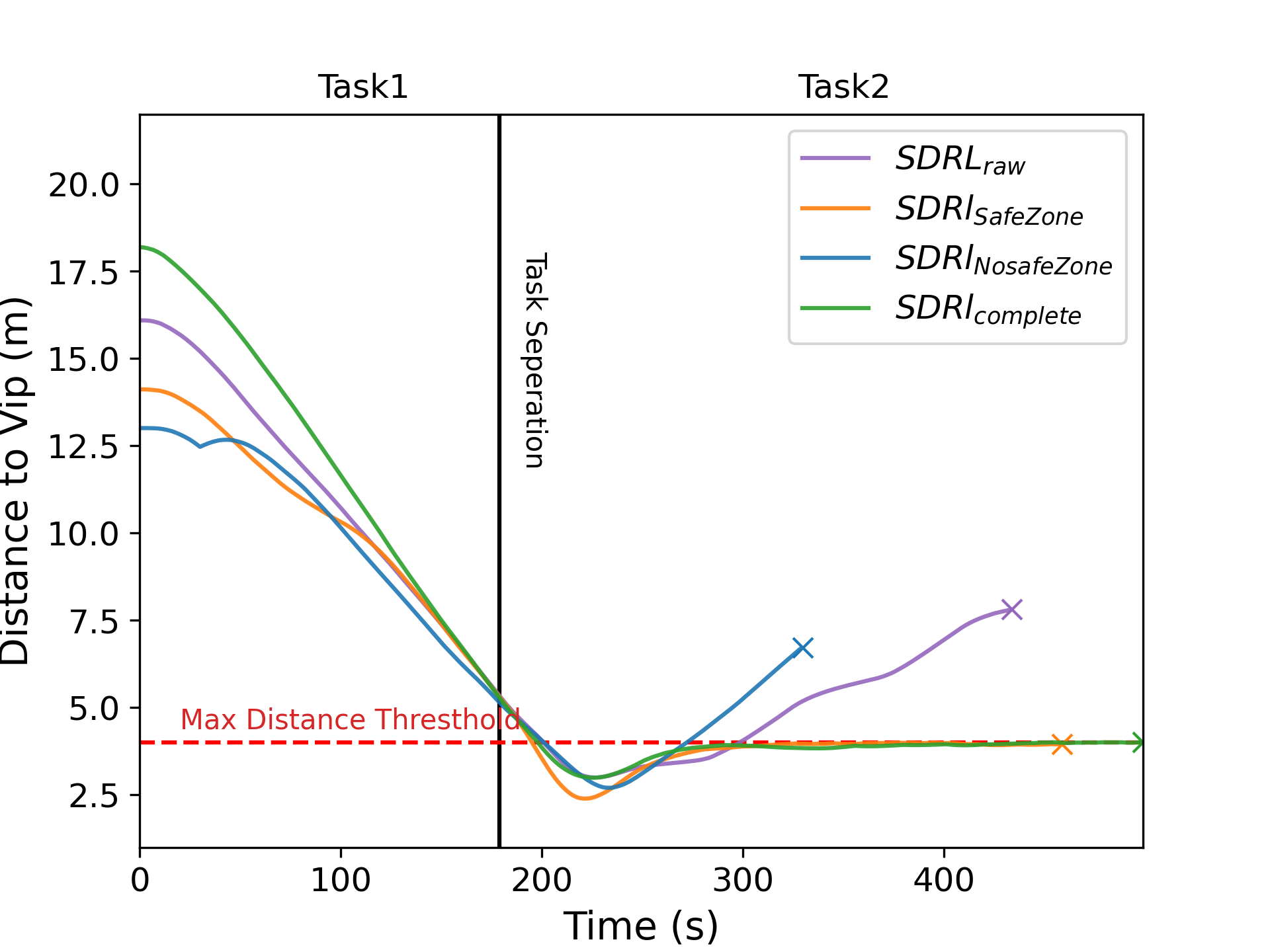}
        
        \label{fig:1}
    \end{subfigure}\hfil 
    \begin{subfigure}{0.33\textwidth }
     \centering 
        \includegraphics[width=\linewidth]{ 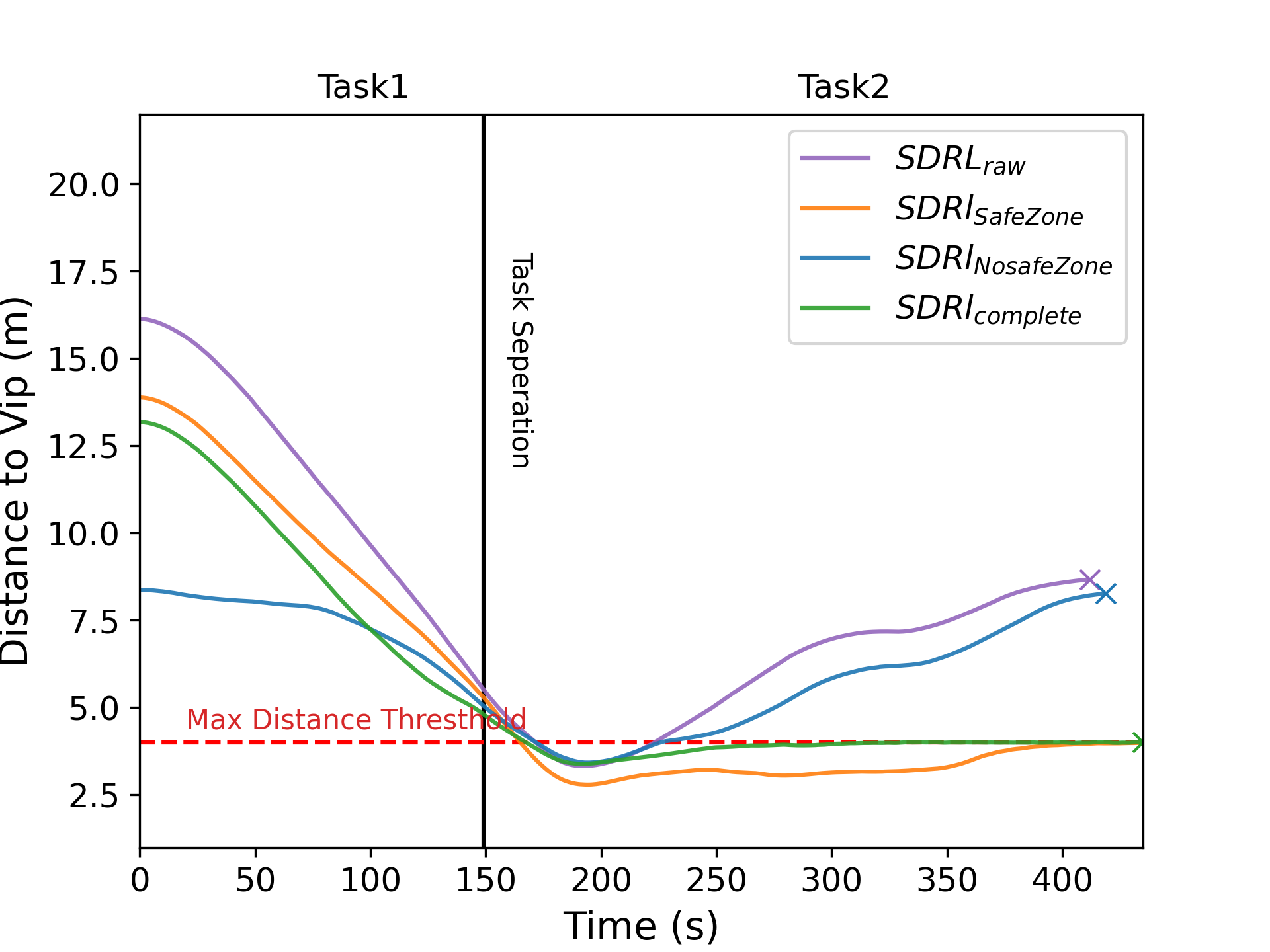}
        
        \label{fig:2}
    \end{subfigure}\hfil 
    \begin{subfigure}{0.33\textwidth}
     \centering 
        \includegraphics[width=\linewidth]{ 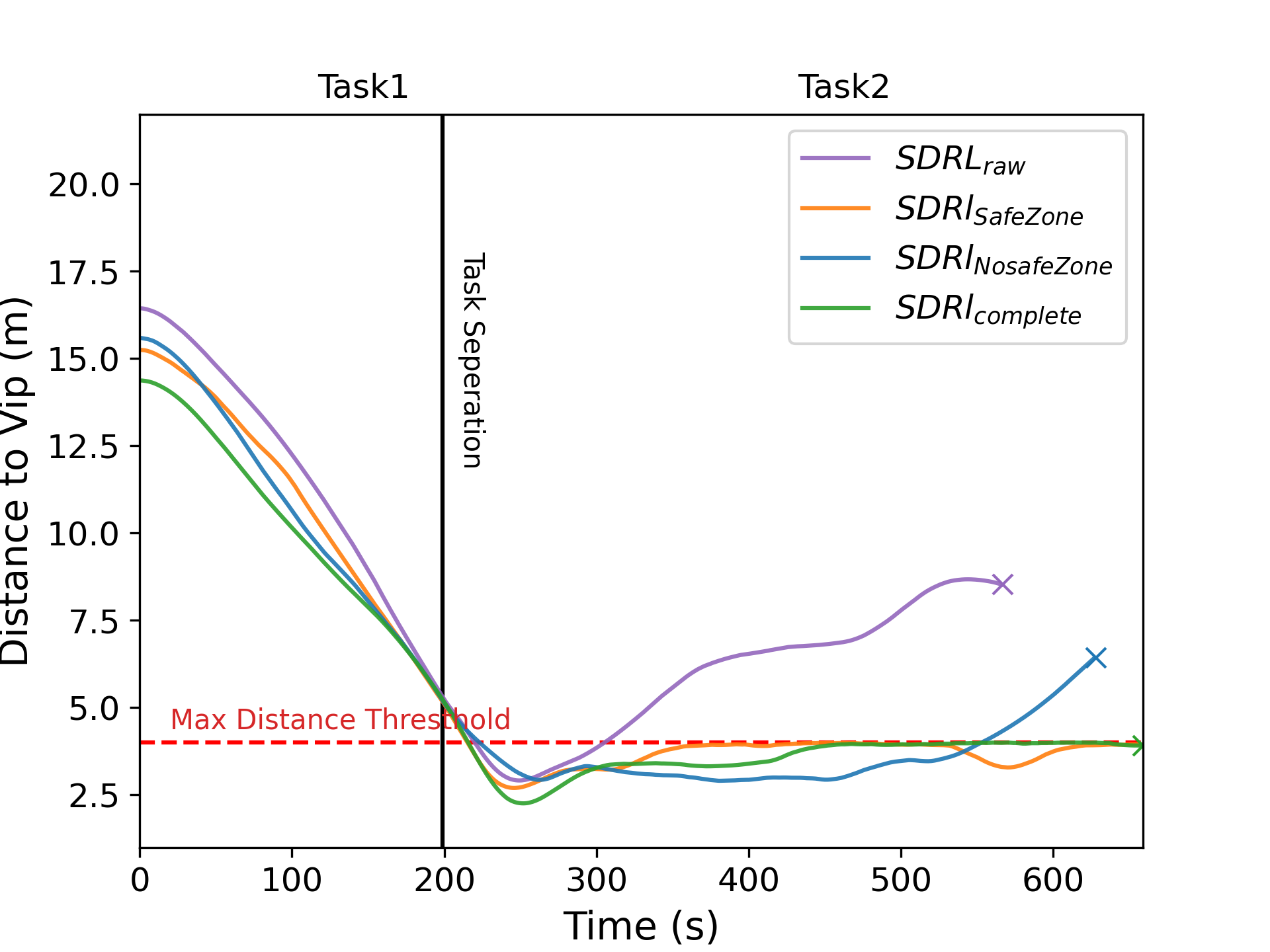}
        
        \label{fig:3}
    \end{subfigure}
        \medskip
    \begin{subfigure}{0.31\textwidth}
         \centering 
        \includegraphics[width=\linewidth]{ 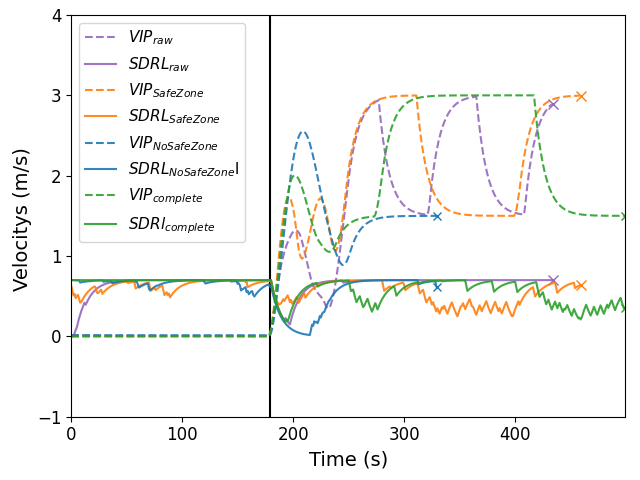}
        
        \label{fig:1}
    \end{subfigure}\hfil 
    \begin{subfigure}{0.31\textwidth }
     \centering 
        \includegraphics[width=\linewidth]{ 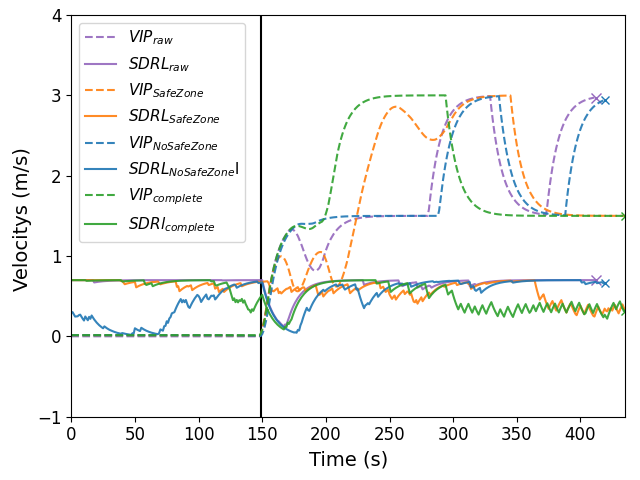}
        
        \label{fig:2}
    \end{subfigure}\hfil 
    \begin{subfigure}{0.31\textwidth}
     \centering 
        \includegraphics[width=\linewidth]{ 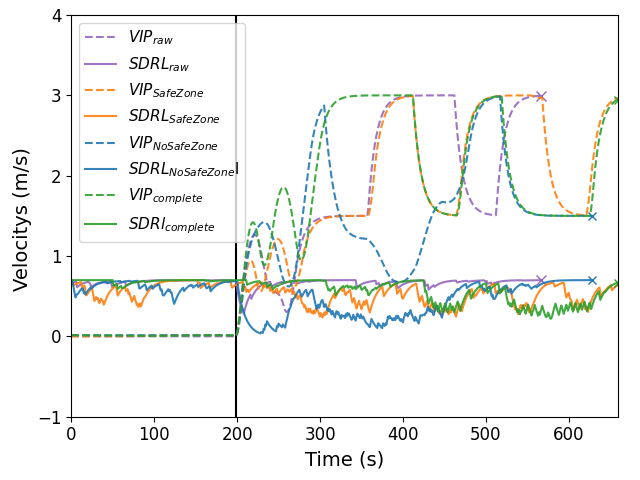}
        
        \label{fig:3}
    \end{subfigure}
    
    \caption{Evaluations of the robot following task: upper row: trajectories of all agents and respective VIP requesting robot following, mid row: the distances of all VIPs and respective agents over the time, lower row: the velocities of all VIPs and respective agents over the time.}
    \label{eval-semantic}
\end{figure*}

\subsection{High Level Tasks}
\noindent After the navigational performance of our semantic agents was demonstrated, we evaluate their ability to carry out the two high-level tasks: human-following, and human-guiding in crowded environments. The results are plotted in Fig. \ref{eval-semantic}.
We recorded the trajectories of the VIP requesting assistance as well as the trajectories of our agents in the upper row of Fig. \ref{eval-semantic}. To demonstrate the ability of our agents to adapt to human behavior, we plotted the distance and velocity of both actors over time in the mid- and lower row of Fig. \ref{eval-semantic}, respectively. The start of an assistance task is indicated in the plot as Task 2. It is observed that the distance of the $SDRL_{complete}$ and the $SDRL_{safe}$ agents are at a constant level, even after switching to Task 2 in all scenarios. Contrarily, the distance of the raw agent $SDRL_{raw}$ and the agent without safety distances increases over time, which occurred in situations where the human got stuck while the robot continued to move. This indicates that our semantic agents are able to adapt to the complex behavior patterns of humans due to the additional information provided to the agent. Similarly, the velocities over time confirm these findings. Whereas the raw agent keeps a constant velocity even when the human's velocity changed, our semantic agent's velocity (orange) adapts well to the human velocity, which is another indication of the semantic agent's ability to adapt its behavior to the human. For instance, in Fig. \ref{eval-semantic} (b) at 270s, the velocity of our semantic agents (green and yellow) increases and decreases synchronously with the VIP velocity, whereas the raw agent (blue) keeps a constant velocity even with decreasing VIP velocity at 300s. 
The results demonstrated that the raw DRL agents without semantic information navigate without paying attention to the human, which is problematic especially in crowded environments as the humans often got stuck while the robot continued to navigate towards the goal. Especially in scenarios with a large number of obstacles, a large discrepancy between robot and VIP velocity and distance is observed.
On the other hand, our semantic agents could adapt their behavior to the human VIP by keeping a constant distance as well as change the velocity according to the human. This enhances human-robot interaction significantly.

\section{Conclusion}
\noindent In this paper, we proposed a DRL-based assistance agent for human assistance in crowded environments. The semantic agent is not only able to navigate safely in highly dynamic environments, but also able to follow or guide a human while adapting its behavior accordingly. We enhanced the agent's observation space with semantic information about the obstacle position, social states, and safety models. Subsequently, we trained the robot to guide or follow a human in crowded environments. Results demonstrated the ability of our agent to adapt its behavior to the human by keeping a constant distance to the target while also adjusting its velocity according to the human. Moreover, the agent was able to navigate safely through crowded environments and accomplished safer and more robust navigation compared to a baseline DRL approach without semantic information. Future works include the incorporation of multi-modal information like sound and visual cues to learn even more complex tasks. Furthermore, we aspire to deploy the approaches towards real robots using computer vision approaches and visual sensors to acquire the semantic information, which in this paper was assumed to be the ground truth data.


\addtolength{\textheight}{-1cm} 




\typeout{}
\bibliographystyle{IEEEtran}
\bibliography{main}

\end{document}